\documentclass[a4paper,fleqn]{cas-dc}

\usepackage[numbers,sort&compress]{natbib}

\usepackage{float}

\usepackage{balance}
\usepackage{threeparttable}

\def\tsc#1{\csdef{#1}{\textsc{\lowercase{#1}}\xspace}}
\tsc{WGM}
\tsc{QE}
\tsc{EP}
\tsc{PMS}
\tsc{BEC}
\tsc{DE}

\begin{document}
\let\ref\Cref 		
\let\eqref\Cref 	
\let\autoref\Cref 	
\let\WriteBookmarks\relax
\def\floatpagepagefraction{1}
\def\textpagefraction{.001}

\footmarks{}

\bookmark[named = FirstPage]{Multi-View Fusion Neural Network for Traffic Demand Prediction} 

\title [mode = title]{Multi-View Fusion Neural Network for Traffic Demand Prediction}    




\author{Dongran Zhang}
\ead{zhangdr5@mail2.sysu.edu.cn}

\author{Jun Li}
\cormark[1]
\cortext[cor1]{Corresponding author}
\ead{stslijun@mail.sysu.edu.cn}

\address{School of Intelligent Systems Engineering, Sun Yat-Sen University, Shenzhen, Guangdong, 518107, China}

\begin{abstract} 
The extraction of spatial-temporal features is a crucial research in transportation studies, and current studies typically use a unified temporal modeling mechanism and fixed spatial graph for this purpose.  However, the fixed spatial graph restricts the extraction of spatial features for similar but not directly connected nodes, while the unified temporal modeling mechanism overlooks the heterogeneity of temporal variation of different nodes. To address these challenges, a multi-view fusion neural network (MVFN) approach is proposed. In this approach, spatial local features are extracted through the use of a graph convolutional network (GCN), and spatial global features are extracted using a cosine re-weighting linear attention mechanism (CLA). The GCN and CLA are combined to create a graph-cosine module (GCM) for the extraction of overall spatial features.  Additionally, the multi-channel separable temporal convolutional network (MSTCN) makes use of a multi-channel temporal convolutional network (MTCN) at each layer to extract unified temporal features, and a separable temporal convolutional network (STCN) to extract independent temporal features. Finally, the spatial-temporal feature data is input into the prediction layer to obtain the final result. The model has been validated on two traffic demand datasets and achieved the best prediction accuracy.

\end{abstract}



\received { 20 February 2023}
\revised { 7 June 2023}
\accepted { 5 June 2023}
\online { 9 June 2023}


\maketitle
\section{Introduction}
\label{SE:Inroduction}

With the rapid development of cities and the application of traffic big data technology, the intelligent transportation system has become a hot research topic, and traffic demand prediction is an important research direction of the intelligent transportation system \cite{ye2021coupled}. Traffic demand and supply in various nodes of cities are often unbalanced \cite{bai2021deep}. Accurately predicting traffic demands in different nodes can help guide taxi drivers to different areas in advance, leading to a re-distribution of transport resources and providing better travel services for residents.

Early research methods are mainly statistical models. Autoregressive integrated moving average (ARIMA) is widely used in traffic flow prediction \cite{1980On}. Kalman filter is also used in traffic prediction\cite{1984Dynamic}. Exponential smoothing and its improved algorithms can find trends, periodicity, and seasonality in time series data \cite{dudek2021hybrid}. The Prophet model decomposes time series into four main components: trend, seasonality, holiday effects, and noise \cite{taylor2018forecasting}. The bimodal Gaussian inhomogeneous Poisson algorithm also reveals the changing trend of bike usage \cite{huang2018bimodal}. Although statistical methods can construct models using well-structured mathematical parameters, they are more suitable for predicting with stable data.

Based on traditional machine learning algorithms, various regression methods are applied to predict data in bigger bike sharing datasets, although the error increases over longer horizons \cite{giot2014predicting}. For instance, gradient boosting decision tree (GBDT) automatically filters the characteristics of variables to achieve traffic generation prediction \cite{li2021urban}. Data are first clustered and filtered through K-means and then public bicycle demands are predicted using support vector machines (SVM) \cite{xu2013public}. The K-Nearest Neighbors (KNN) algorithm is also used for multi-step prediction \cite{yu2016k}. These methods can capture certain temporal features in nonlinear traffic data, but cannot obtain spatial features.

The rapid development of deep learning has provided new methods for spatial-temporal prediction. In terms of spatial features extraction, convolutional neural network (CNN) \cite{song2017traffic} and graph convolutional network (GCN) \cite{kipf2016semi} are frequently used to extract adjacent spatial feature of spatial units represented by grids and nodes in graphs, respectively, and distant spatial features can be extracted by adaptive matrix generation \cite{wu2019graph, ye2021coupled} and attention mechanism \cite{vaswani2017attention, guo2021learning}. In terms of temporal feature extraction, models including long short-term memory network (LSTM) \cite{hochreiter1997long}, gated recurrent unit network (GRU) \cite{cho2014learning}, and temporal convolutional network (TCN) have been introduced \cite{bai2018empirical}. By combining spatial and temporal extraction modules, deep learning models have achieved excellent results in various tasks of spatial-temporal prediction.

Traffic demand prediction is a complex spatial-temporal problem. Although high accuracies have been achieved through various deep learning models, there are still two challenges in extracting spatial and temporal features:
\begin{itemize}
\item Spatial features include local and global features. The demand is influenced by neighboring nodes and will show strong correlations locally. Furthermore, nodes with similar functions will generate similar demand changes, and these nodes are generally not directly connected. It is crucial to implement the fusion of local and global features to effectively extract complete spatial features, thereby improving the accuracy of prediction.

\item Temporal features include unified and independent features. The traffic demand of various nodes is heterogeneous, and each node shows two modes of pick-up and drop-off. Using a unified module to extract temporal features alone may lead to ignorance of the heterogeneity of the temporal variation in each node. Therefore, additional research is essential to determine how to achieve independent temporal feature extraction and integrate it with unified temporal feature extraction while taking into account the heterogeneity of temporal features.
\end{itemize}

To address the challenges mentioned earlier, a multi-view fusion neural network (MVFN) model is proposed for traffic demand prediction. In the spatial dimension, the proposed model extracts and fuses spatial features from both local and global views. In the temporal dimension, the model extracts and fuses temporal features from both unified and independent views.

The main contributions of this paper are as follows.
\begin{itemize}
\item In the spatial dimension, a fusion method of local and global features is proposed. First, a graph convolutional network module (GCN) is used to obtain local features. Then, a cosine re-weighting linear attention module (CLA) is used to efficiently obtain global features. Finally, the graph-cosine module (GCM) is formed by combining GCN and CLA to achieve the fusion of spatial local and global features.

\item In the temporal dimension, a multi-channel separable temporal convolutional network (MSTCN) is proposed to extract temporal features. In each layer of MSTCN, unified temporal features are extracted using a multi-channel temporal convolutional network (MTCN) and independent temporal features using a separable temporal convolutional network (STCN), while the use of a dilated causal convolution operation increases the receptive field.

\item Extensive experiments are conducted on two real-world datasets, and the experimental results show that the performance of the model in this paper consistently outperforms than the existing related prediction methods.
\end{itemize}

\section{Related Work}
\label{se:Related Work}

\subsection{Deep learning for traffic prediction}
\label{se:Deep learning for traffic prediction}

As a popular research direction, deep learning methods have gained increasing popularity in traffic spatial-temporal prediction. Various models and methods have been proposed for extracting spatial-temporal features.

In the feature extraction of spatial dimensions, the mainstream methods are based on CNN, GCN \cite{zhuo2024partitioning}, and attention mechanism. Transportation researchers divide spatial units into regular grid data, treat traffic data as image-like data, and use CNN to extract spatial proximity correlations, which are well-suited for regular traffic data prediction \cite{song2017traffic}. But on the other hand, the analysis units in traffic data are more similar to the nodes on the graph, so the GCN method can be used to process the traffic spatial data, and the information of the neighbor nodes can be aggregated on the fixed traffic graph. For example, DCRNN uses bidirectional diffusion random walk on the graph to capture spatial features \cite{li2017diffusion}. Graph WaveNet adds a self-adaptive adjacency matrix to enrich the extracted spatial features \cite{wu2019graph}. CCRNN adopts singular value decomposition for learning different convolutional layers and employs coupled layer-wise mechanism for overall spatial feature modeling \cite{ye2021coupled}. MDRGCN combines similarity matrix and adaptive matrix to fuse spatial features of different patterns \cite{huang2022multi}. TGANet uses Jaccard distance, Pearson correlation coefficient and attention mechanism to jointly model spatial features \cite{chen2022traffic}. MVDGCN considers weekly and daily patterns on the basis of CCRNN \cite{huang2023multi}. GCN-DHSTNet adopts GCN modeling to include spatial features of recent, daily, and weekly volume \cite{ali2022exploiting}. PKET-GCN utilizes the prior knowledge extraction module for spatial feature modeling, including the static node feature matrix, the node correlation matrix constructed with Pearson correlation coefficients, and the environmental measurement matrix \cite{bao2023pket}. DDGCRN applies spatial-temporal embedding generation and node adaptive parameter learning to model spatial features \cite{weng2023decomposition}. DTC-STGCN adopts GCN module based on dynamic adjacency matrix and attention mechanism to learn dynamic spatial features \cite{xu2023dynamic}. MISTAGCN combines spatial attention and GCN to model spatial features \cite{tao2023multiple}. STSSN considers node position embedding and time step embedding, and adopts enhanced diffusion convolution network to model spatial features \cite{cao2022spatio}. ASTGNN uses GCN and self attention mechanism multiplication to obtain a dynamic graph convolutional network, which takes into account the similarity of adjacent nodes in input data \cite{guo2021learning}. MGT utilizes meta-learning to obtain spatial-temporal embedding and constructs a transition matrix based on three matrices: distance, functional similarity, and OD \cite{ye2022meta}. 

In the feature extraction of temporal dimensions, on the other hand, the mainstream methods are based on recurrent neural network (RNN), TCN, and attention mechanism. In the research of temporal features extraction based on RNN, LSTM leverages input gate, forget gate, and output gate to control input values, memory values, and output values, effectively solving the gradient vanishing and exploding problems of RNN in long sequence prediction, and is frequently used in temporal feature modeling \cite{2019STG2Seq, ali2022exploiting, xu2023dynamic}. GRU adopts update gates and reset gates, which simplifies the calculation of LSTM while achieving comparable results, making it widely used in temporal feature modeling \cite{li2017diffusion, ye2021coupled, huang2022multi,weng2023decomposition, huang2023multi}. However, LSTM and GRU cannot be calculated in parallel, reducing the computational speed. Therefore, TCN composed of dilated causal convolution is also applied in temporal feature modeling, significantly reducing the number of parameters while ensuring prediction accuracy \cite{STGCN, wu2019graph, cao2022spatio}. The temporal attention mechanism can adaptively simulate nonlinear correlations between different time steps \cite{zheng2020gman, guo2021learning, ye2022meta, tao2023multiple}.

\subsection{Simplified computation of attention mechanism}
\label{se:Simplified computation of attention mechanism}
The attention mechanism focuses on the more critical information and reduces the attention to irrelevant information through a weighting mechanism \cite{vaswani2017attention}. Specifically, attention mechanisms have been successfully employed in both spatial and temporal dimensions for dynamic modeling and prediction of traffic spatial-temporal data, leading to excellent performances \cite{zheng2020gman, guo2021learning, ye2022meta}. 

The attention mechanism, which can achieve information transmission distance of $O(1)$ between any two nodes, can also be viewed as a fully connected graph \cite{kreuzer2021rethinking}. However, as operations between any two nodes are required, runtime complexity increases dramatically in a quadratic manner $O(N^2)$ as the number of nodes increases, resulting in inefficient computation. As a result, many studies aim to reduce the computational complexity \cite{tay2020efficient}. For instance, some scholars have used low-rank matrix approximation methods to decrease computation time, as the attention matrix is observed to be low-rank \cite{lu2021soft}. Additionally, a pre-defined pattern can be used to limit the field view of the self-attention mechanism through sparsification techniques \cite{child2019generating}. Sparse calculation methods can also be obtained from the perspective of data distribution \cite{zhou2021informer, kitaev2019reformer}.

\section{Preliminaries}
\label{se:Preliminaries}

\subsection{Graph structure}
\label{se:Graph structure}
The traffic demand graph can be represented by considering each region in the transportation system as a node. With $ G = (V, E, A) $, $V$ represents the set of nodes in the graph, $ E$ represents the set of node edges, and $ A$ represents the spatial connection of nodes. The adjacency matrix is  $ A\in \mathbb{R}^{N\times N} $. At time step $ t $, the feature matrix of the graph $ X_{t} \in \mathbb{R}^{N\times f} $, where $ f $ refers to the feature dimension of the input, and $ N$ refers to the number of nodes.  

\begin{equation}
	\label{eq:Aij}
	A_{ij} =\left\{\begin{matrix}
		1, & (v_{i},v_{j}) \in E\\
		0,& otherwise
	\end{matrix}\right.
\end{equation}

\subsection{Traffic demand prediction}
\label{se:Traffic demand prediction}
Traffic demand prediction poses the problem of learning from historical traffic graph signal data with $P$ time steps to obtain a mapping function $\Gamma$ for predicting future traffic graph signal data over $Q$ time steps.

\begin{equation}
	\label{eq:prediction}
	X_{t+1:t+Q}= \Gamma \left [ X_{t-P+1:t}, G\right ] 
\end{equation}
where $ \Gamma$ is the nonlinear mapping function to be learned, $ X_{t+1:t+Q} \in \mathbb{R}^{Q\times N \times f}$, $ X_{t-P+1:t} \in \mathbb{R}^{P\times N \times f}$. 

\section{Methodology}
\label{se:Methodology}
\begin{figure*}
	\centering
	\includegraphics[width=7.0in]{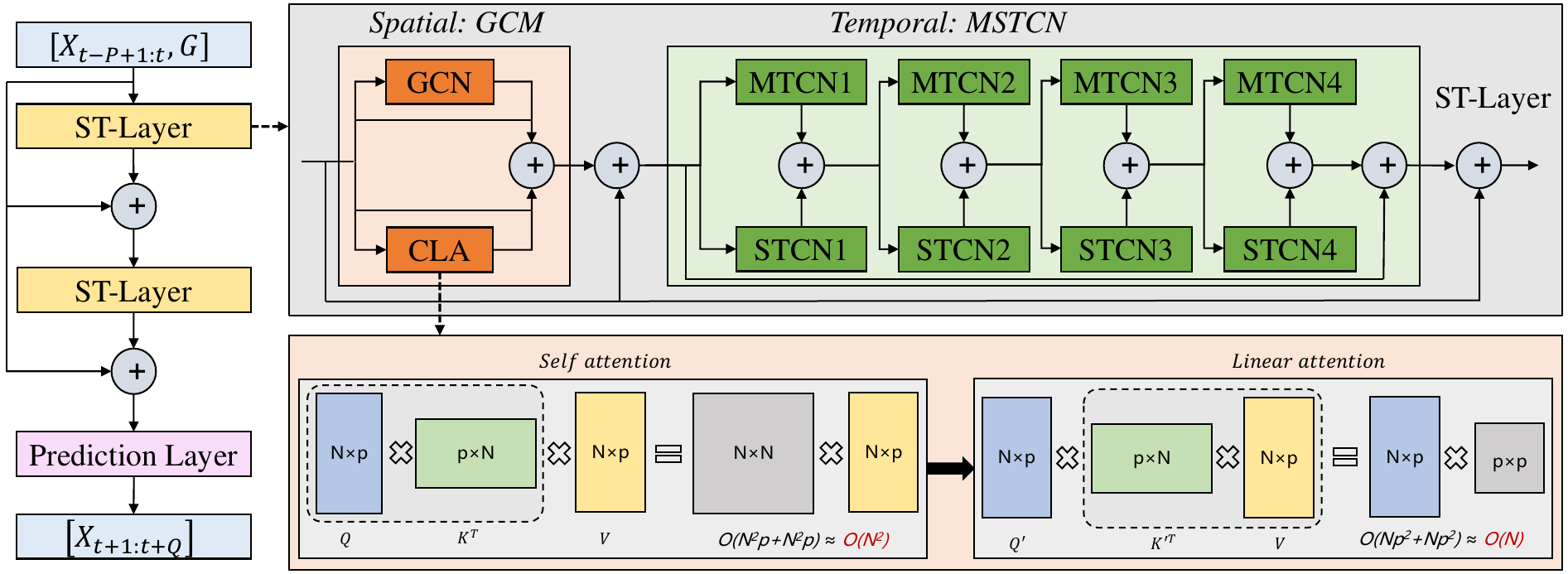}
	\caption{The overall structure of the MVFN model.}
	\label{fig:Model framework}
\end{figure*}

The overall structure of the MVFN is shown on the left side of \ref{fig:Model framework}, which includes two ST-Layers, one prediction layer, and several residual connections. The upper right side of \ref{fig:Model framework} shows the internal structure of each ST-Layer, including the spatial features extraction module GCM and the temporal features extraction module MSTCN. GCM includes a spatial local features extraction submodule GCN and a spatial global features extraction submodule CLA. As shown in the bottom right of \ref{fig:Model framework}, linear attention is achieved by adjusting the dot product order in CLA. Additionally, MSTCN has a total of four layers that are stacked, with each layer containing a temporal unified features extraction module MTCN and a temporal independent features extraction module STCN. 

\subsection{Spatial local features extraction module}
\label{se:Spatial local features extraction module}
The traffic demand of a node is influenced by neighboring nodes. In the extraction of spatial local features, we use GCN to learn the structural features of topological graphs, fuse regional information with neighboring information, and capture spatial local dependencies on the graph \cite{kipf2016semi,  zhuo2022efficient}. Consider a multi-layer GCN, where the propagation of the $ l$-th layer can be written as:

\begin{equation}
	\label{eq:H}
	H^{(l+1)}= {\rm ReLU} (\tilde{D}^{-\frac{1}{2}}  \tilde{A}\tilde{D}^{-\frac{1}{2} }H^{(l)}W^{(l)})
\end{equation}
where $ \tilde{A} = A+I_{N}$, $ A$ is the adjacency matrix, $ I_{N} $ is self-connection; $ \tilde{D}_{ii} =\sum_{j}\tilde{A}_{ij}$, $ \tilde{D}$ is a diagonal matrix, where each element on the diagonal represents the degree of the corresponding point in the graph; $ H^{(l)}$ and $ H^{(l+1)}$ are the input and output of the $ l$-th layer, respectively; $ W^{(l)}$ is the learnable parameter of $ l$-th layer; and ReLU is the activation function.

\subsection{Spatial global features extraction module}
\label{se:Spatial global features extraction module}
Spatial global features extraction via self-attention can reveal similar trends in traffic demand patterns of non-adjacent but functionally similar nodes \cite{ye2022meta}. While sparse attention mechanisms can simplify computational complexity, with only a subset of nodes in the calculation, they contradict the goal of extracting spatial global features in this study. Therefore, linearization optimization methods are more suitable for our task. To extract spatial global features, we adopt the cosine re-weighting linear attention mechanism (CLA) of CosFormer \cite{qin2021cosformer}. CLA, as presented in the lower right side of \ref{fig:Model framework}, enables linearization calculation by modifying the dot product order and utilizes the cosine re-weighting method to reinforce local correlation.

In the calculation of the standard self-attention mechanism, the input data $ X$ is first mapped to $ Q$, $ K$, $ V$ through linear transformation, with the formulas below:

\begin{align}
	&Q, K, V = XW_{Q}, XW_{K}, XW_{V} \\
	&S(Q_{i},K_{j}  )=\exp^{Q_{i}K_{j}} \\
	& A_{i} = {\textstyle \sum_{j}^{}} \frac{S(Q_{i},K_{j})}{ {\textstyle \sum_{j}^{}S(Q_{i},K_{j})}} V_{j}
	\label{eq:attention}
\end{align}
where $ W_{Q}$, $ W_{K}$, $ W_{V}$ are three different learnable parameters. $ S (\cdot ) $ is the dot product similarity. $ A_{i}$ is the value of the attention matrix in the $ i$-th row.

By converting the attention mechanism to a linear calculation method and replacing the similarity measure with a ReLU function, the computational complexity is significantly reduced. Specifically, with $ p = P \times f$, and given that $ N >> p$ in experiments, the computational complexity can be reduced from  $ O(N^{2})$ to $ O(N)$.

\begin{align}
	&LS(Q_{i},K_{j})=\phi(Q_{i})\phi(K_{j})^{T} \\
	&LA_{i} =\frac{ {\textstyle \sum_{j}\phi(Q_{i})\phi(K_{j})^{T}}V_{j}}{ {\textstyle \sum_{j}\phi(Q_{i})\phi(K_{j})^{T}}} = \frac{ {\textstyle \phi(Q_{i})\sum_{j}\phi(K_{j})^{T}}V_{j}}{ {\textstyle \phi(Q_{i})\sum_{j}\phi(K_{j})^{T}}} 
	\label{eq:LAI}
\end{align}
where $ \phi (\cdot ) $ is the value mapped by the ReLU, $ LS(\cdot)$is a computational function that can be split linearly, and $ LA_{i}$ is the attention value of the $ i$-th row after multiplicative swapping.

The introduction of the cosine re-weighting mechanism incorporates the relative position deviation into the attention matrix, thus enhancing locality, which aligns with the spatial features distribution of traffic demand and stabilizes the training process. To simplify the representation, $Q'$ denotes $\phi(Q)$, and $K'$ denotes $\phi(K)$. The expression of the formula is given below:

\begin{equation}
	\label{eq:cosin}
	CLS(Q_{i}^{'},K_{j}^{'} )=Q_{i}^{'}K_{j}^{'T}\cos(\frac{\pi }{2} \times \frac{i-j}{N} ) 
\end{equation}
where $ CLS(\cdot)$ is the cosine re-weighting function, and expand the above formula to get:

\begin{equation}
	\begin{split}
		\label{eq:expend cosin}
		CLS\left (Q_{i}^{'}, K_{j}^{'} \right ) & = \left (Q_{i}^{'} \cos\left ( \frac{\pi i}{2N} \right )   \right ) \left (K_{j}^{'} \cos\left ( \frac{\pi j}{2N} \right )   \right ) ^{T} \\ 
		& + \left (Q_{i}^{'} \sin\left ( \frac{\pi i}{2N} \right )   \right ) \left (K_{j}^{'} \sin\left ( \frac{\pi j}{2N} \right )   \right ) ^{T}
	\end{split}
\end{equation}

Each part of the above formula is abbreviated as follows:

\begin{equation}
	\begin{aligned}
		\label{eq:simple cosin}
		& Q_{i}^{\cos}, K_{j}^{\cos}=Q_{i}^{'} \cos\left ( \frac{\pi i}{2N} \right ), K_{j}^{'} \cos\left ( \frac{\pi j}{2N} \right ) \\
		& Q_{i}^{\sin}, K_{j}^{\sin}=Q_{i}^{'} \sin\left ( \frac{\pi i}{2N} \right ), K_{j}^{'} \sin\left ( \frac{\pi j}{2N} \right ) \\
	\end{aligned}
\end{equation}

We substitute $ CLS(\cdot)$ into Equation \ref{eq:attention} to obtain linearized attention after cosine re-weighting . The formula is as follows:

\begin{equation}
	\label{eq:final cosformer}
	CLA_{i} =\frac{\sum_{j=1}^{N}Q_{i}^{\cos}\left ( \left(K_{j}^{\cos}\right )^{T} V_{j}  \right )+\sum_{j=1}^{N}Q_{i}^{\sin} \left (  \left ( K_{j}^{\sin}  \right )^{T}V_{j} \right )  }
	{\sum_{j=1}^{N}Q_{i}^{\cos}\left ( K_{j}^{\cos}  \right )^{T}+ \sum_{j=1}^{N}Q_{i}^{\sin} \left ( K_{j}^{\sin}  \right )^{T}     } 
\end{equation}
where $ CLA_{i}$ refers to the attention value of the $ i$-th row. Finally, the $ CLA$ formula obtained without losing the generality is as follows:

\begin{equation}
	\label{eq:CLA}
	CLA = Q^{\cos}(K^{\cos}V) + Q^{\sin}(K^{\sin}V)
\end{equation}

\subsection{Multi-channel separable temporal convolutional network }
\label{se:MSTCN}

The overall structure of MSTCN is shown in the light green section of the ST-Layer on the upper right side of \ref{fig:Model framework}. It is stacked in four layers, each layer containing an MTCN module and an STCN module, facilitating the extraction of unified and independent temporal features, respectively.

\begin{figure}
	\centering
	\includegraphics[width=3.0in]{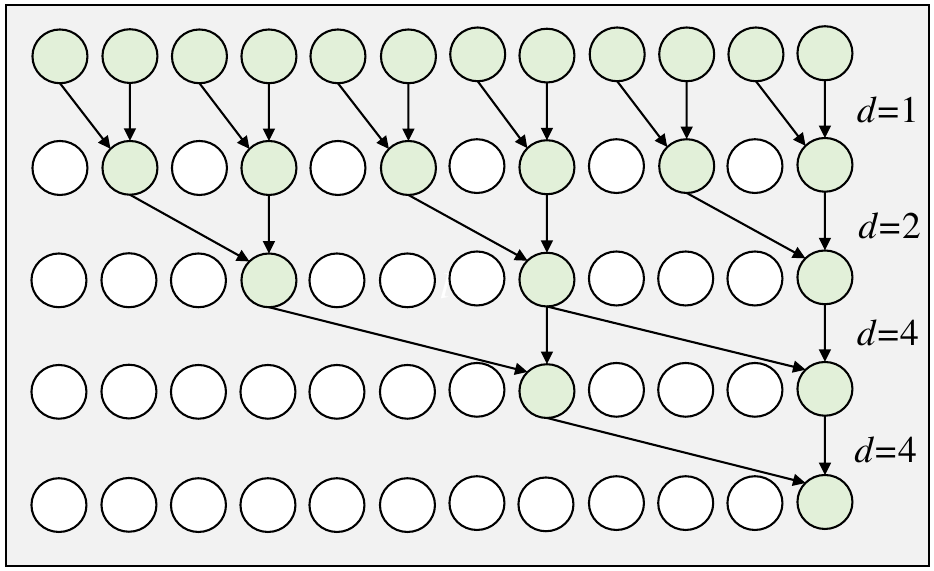}
	\caption{Dilated causal convolution. The kernel size is 2, and the dilation coefficients $ d$ are 1, 2, 4, and 4, respectively.}
	\label{fig:Tcn}
\end{figure}

MTCN and STCN both rely on dilated causal convolution \cite{bai2018empirical}. As shown in Figure~\ref{fig:Tcn}, dilated causal convolution utilizes the dilation operation to greatly enhance the receptive field without increasing computational burden, and the causal convolution ensures that only prior data values are considered. In light of the typical period $ P$ of 12 in spatial-temporal prediction tasks, we set the kernel size to 2 and the dilation coefficients to 1, 2, 4, and 4, respectively. Subsequently, after four layers of convolutional operations, the size of the receptive field becomes 12, covering input data features in the temporal dimension. The formula is as follows:

\begin{equation}
	\label{eq:Ds}
	D(s)=\sum_{i=0}^{k-1} f(i)\cdot X_{s-d\cdot i}  
\end{equation}
where $ X$ is the input data, $ d$ is the dilation coefficient, $ k$ is the kernel size, and $ s-d\cdot i$ determines to only perform convolution operations on past input data.

The standard convolution is to use multiple convolution kernels to convolve and extract features on all the input channels, while the grouped convolution groups the channels, and the correlation between groups does not affect each other, resulting in a separate output for each group \cite{xie2017aggregated, howard2017mobilenets}. Given the properties of grouped convolutions, we can use them to construct MTCN and STCN. The comparison of parameter settings between MTCN and STCN is shown in \ref{tab:MTCN and STCN Paremeter}, in which $ N$ is the node of the research data, and $ f$ is the number of feature contained in the node. A brief description of both the MTCN and STCN modules is given below.

\begin{table}[]
	\centering
	\caption{Comparison of parameter settings between MTCN and STCN.}
	\begin{tabular}{@{}cccc@{}}
		\toprule
		\textbf{Parameter} & \textbf{MTCN}  & \textbf{STCN} \\ \midrule
		Input Channel      & $ N \times f$  & $ N \times f$           \\
		Output Channel     & $ N \times f$  & $ N \times f$           \\
		Group              & 1              & $ N \times f$           \\
		Filter Number      & $ N \times f$  & $ N \times f$          \\
		kernel in Filter   & $ N \times f$  & 1             \\ \bottomrule
	\end{tabular}
	\label{tab:MTCN and STCN Paremeter}
\end{table}

\begin{figure}
	\centering
	\subfloat[MTCN]{
		\includegraphics[width=0.45\linewidth]{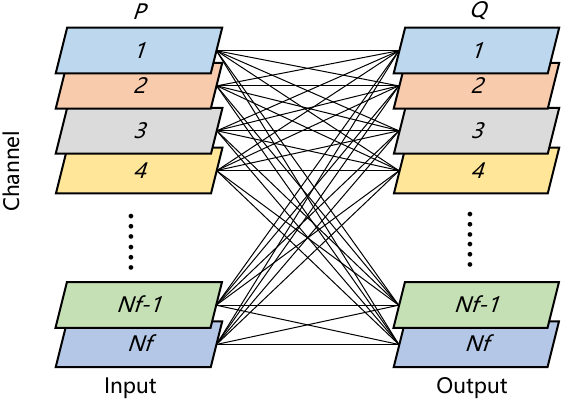}
		\label{fig:MTCN}}
	\subfloat[STCN]{
		\includegraphics[width=0.45\linewidth]{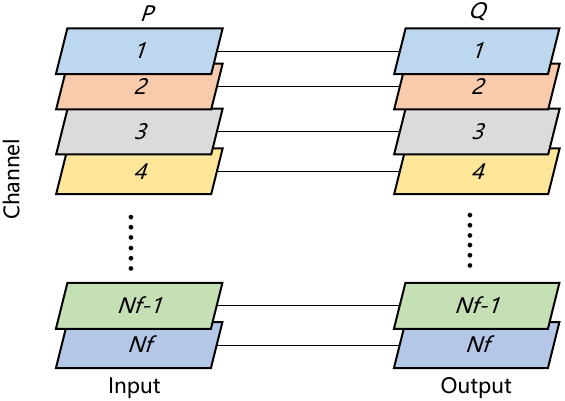}
		\label{fig:STCN}}
	\label{fig:MSTCN}
	\caption{Comparison of two TCNs. (a) MTCN, the output channels are mutually operated with other channels; (b) STCN, each channel operates independently.}
\end{figure}

\textbf{ MTCN: } MTCN is utilized to extract unified temporal features, as illustrated in \ref{fig:MTCN}, the number of grouped convolutions is set to 1, and each output channel is the result of the mutual operation of all input channels, realizing the overall extraction of temporal features. The traffic demand data can be regarded as a set of one-dimensional time series. The input data format is $ P \times N \times f$, and the output data format is $ Q \times N \times f$. In our temporal features extraction, we always keep the size of input channel $ C_{i}$ and the size of output channel $ C_{o}$ equal to $ N \times f $, and the size of  overall convolution kernel is: $ C_{o} \times C_{i} \times k$.

\textbf{ STCN: } STCN is utilized to ensure the independent extraction of temporal features among different channels, as illustrated in \ref{fig:STCN}. STCN is an extremely grouped convolution, which uses the same number of groups as $ C_{i}$. We set the number of $ C_{i}$, $ C_{o}$ and groups to $ N \times f$, the input data formats is $ P \times N \times f$, and output data formats is $ Q \times N \times f$. Different channels are assigned to their respective groups and there is no information exchange between channels, so the independent extraction of temporal features is achieved.

\subsection{Prediction layer}
\label{se:Predict layer}
The prediction layer is composed of two linear transformation layers and an activation function. The data is subject to a linear transformation and the ReLU activation function, and then input into the second linear transformation layer \cite{vaswani2017attention}. The formula is as follows:

\begin{equation}
	\label{eq:Predict layer}
	{\rm PL} (X)= {\rm ReLU}(X W_{1} + b_{1})W_{2} +b_{2} 
\end{equation}
where $ W_{1}$ and $ W_{2}$ are weights, and $ b_{1}$ and $ b_{2}$ are biases. $ X$ represents the input of prediction layer.

\subsection{Loss function}
\label{se:Loss function}
Mean absolute error (MAE) has been selected as the loss function for network training in this paper. The formula is as follows:

\begin{equation}
	MAE=\frac{1}{n}\sum_{i=1}^{n}\left|y_{i}-y_{i}^{p}\right|
	\label{eq:mae}
\end{equation}
where $ n$ is the total amount of data samples, $ y_{i}$ is the ground truth, $ y_{i}^{p}$ is the predicted value of the i-th sample.

\section{Experiments}
\label{se: Experiments}

\subsection{Datasets}
The performance of MVFN is validated on two publicly available traffic datasets from NYC OpenData. The datasets included records for both taxi and bike orders \cite{ye2021coupled}. 

\begin{itemize}
	\item \textbf{NYC Citi Bike}: Bike trip datasets from April 1st, 2016 to June 30th, 2016, including pick-up node, drop-off node, pick-up time, and drop-off time.
	\item \textbf{NYC Taxi}: Taxi datasets from April 1st, 2016 to June 30th, 2016, including pick-up time, drop-off time, pick-up latitude and longitude, drop-off latitude and longitude, and trip distance.
\end{itemize}

\subsection{Baseline methods}
The following models were compared to MVFN:

\begin{itemize}
	\item \textbf{HA}: We calculate the mean of the previous time steps as the current prediction.
	\item \textbf{XGBoost}: XGBoost is a boosted tree algorithm that supports multiple weak learners \cite{chen2016xgboost}.
	\item \textbf{FC-LSTM}: Long short-term memory network is a recurrent neural network with fully connected LSTM hidden units \cite{hochreiter1997long}.
	\item \textbf{DCRNN}: DCRNN employs diffusion graph convolution and GRU to encode spatial and temporal features, respectively \cite{li2017diffusion}.
	\item \textbf{STGCN}: STGCN uses graph chebnet convolution and time gate convolution to form spatial-temporal blocks for spatial-temporal features extraction \cite{STGCN}.
	\item \textbf{STG2Seq}: STG2Seq adopts graph convolutional based on seq2seq structure to predict multi-step traffic demand, and has an attention mechanism module for dynamic modeling \cite{2019STG2Seq}.
	\item \textbf{Graph WaveNet}: Graph WaveNet uses graph convolution to capture spatial dependencies and dynamic causal convolution to capture temporal dependencies \cite{wu2019graph}.
	\item \textbf{CCRNN}: CCRNN adaptively learns different graph adjacency matrices in different layers, and then uses a coupling mechanism to associate matrices at different levels, and predicts the final result with a GRU \cite{ye2021coupled}.
\end{itemize}

\begin{table*}
	\centering
	\renewcommand{\arraystretch}{1.2}
	\caption{Evaluations of MVFN and baselines.}
	\begin{tabular}{ccccccc}
		\hline
		\multirow{2}{*} {Method} & \multicolumn{3}{c}{NYC Citi Bike} & \multicolumn{3}{c}{NYC Taxi} \\
		\cline{2-7}
				     & RMSE & MAE & PCC & RMSE & MAE & PCC  \\
		\hline
		HA             & 5.2003 & 3.4617 & 0.1669 & 29.7806 & 16.1509 & 0.6339 \\
		XGBoost        & 4.0494 & 2.4690 & 0.4861 & 21.1994 & 11.6806 & 0.8077 \\
		FC-LSTM        & 3.8139 & 2.302  & 0.5675 & 18.0708 & 10.2200 & 0.8645 \\
		DCRNN          & 3.2094 & 1.8954 & 0.7227 & 14.7926 & 8.4274  & 0.9122 \\
		STGCN          & 3.6042 & 2.760  & 0.7316 & 22.6489 & 18.4551 & 0.9156 \\
		STG2Seq        & 3.9843 & 2.4976 & 0.5152 & 18.0450 & 9.9415  & 0.8650 \\
		Graph WaveNet  & 3.2943 & 1.9911 & 0.7003 & 13.0729 & 8.1037  & 0.9322 \\
		CCRNN          & 2.8382 & 1.7404 & 0.7934 & 9.5631  & 5.4979  & 0.9648 \\
		\textbf{MVFN}  &\textbf{2.6857} &\textbf{1.6694} &\textbf{0.8119} &\textbf{9.1617} &\textbf{5.1726} &\textbf{0.9667} \\
		\hline
	\end{tabular}%
	\label{tab:Evaluations}%
\end{table*}%

\subsection{Experiment settings}
In the two datasets, the NYC Citi Bike contains 250 virtual nodes, and the NYC Taxi contains 266 virtual nodes, but each node contains two features, i.e. pick-up and drop-off. The time interval between data records is 30 minutes. In the last four weeks of the data sample, the first two weeks are used as the validation datasets and the last two weeks are used as the test datasets, the remaining data are used for model training.

$P$ and $Q$ are both set to 12, and the model is entirely trained on the training datasets. When extracting the spatial local features, we stack the two types of features together, and the data format is $ (P \times f) \times N$, which is extracted for 250 and 266 nodes respectively. When extracting the spatial global features, we multiply spatial nodes with features as the total global node, which enables flexible interaction of different features, and the data format is  $ P \times (N \times f)$, which is extracted for 500 and 532 nodes respectively. When extracting temporal features, the data processing method is the same as extracting spatial global features, and the data format is  $ P \times (N \times f)$. The number of input and output channels is equal to the number of nodes, which are 500 and 532, respectively. The Pytorch framework is used to build the model. The training epoch is 100. The learning rate is 0.001. The batch size is set to 64. The Adam algorithm is used for optimization. 

Three metrics are employed to evaluate the model performance, namely root mean squared error (RMSE), MAE \ref{eq:mae}, and Pearson correlation coefficient (PCC).
\begin{align}
	&RMSE=\sqrt{\frac{1}{n}\sum_{i=1}^{n}(y_{i}-y_{i}^{p})^{2}} \\
	&PCC=\frac{ {\textstyle \sum_{i=1}^{n}(y_{i}^{p}-\bar{y^{p}})(y_{i}-\bar{y})}}{\sqrt{\sum_{i=1}^{n}(y_{i}^{p}-\bar{y^{p}})^{2}}\sqrt{\sum_{i=1}^{n}(y_{i}-\bar{y})^{2}} } 
	\label{eq:PCC}
\end{align}
where $ n$ is the total number of data, $ y_{i}$ is the ground truth, $ y_{i}^{p}$ is the predicted value, $ \bar{y}$ is the average of the ground truth, and $ \bar{y^{p}}$ is the average of the predictions.

\subsection{Experiment results}
\ref{tab:Evaluations} shows the comparison of different models.  The model is trained 10 times and outputs are averaged as the results, which show that MVFN consistently outperforms the baseline model on both datasets.

HA, XGBoost, and FC-LSTM can only take into account temporal features and thus perform poorly in large-scale spatial-temporal data prediction. Other baseline algorithms can capture spatial and temporal features at different levels, thus improving the accuracy of predictions. 

MVFN fuses spatial-temporal features, local and global spatial features, and unified and independent temporal features, thus achieving the best prediction results. CCRNN is a competitive model and as a comparison, on the  NYC Citi Bike, the RMSE of our model decreases by 5.37\%, the MAE decreases by 4.08\%, and the PCC increases by 2.33\%. On the NYC Taxi, the RMSE of our model decreases by 4.2\%, MAE decreases by 5.92\% and PCC increases by 0.19\%.

\subsection{Ablation study}

To verify the effectiveness of each module, we conducted a series of experiments with fixed random seeds. We set up 6 ablation comparison experiments, which are: NO-GCN: the spatial features extraction module of GCN is removed, that is, there is no local connection; NO-CLA: the CLA global spatial features extraction module is removed, there is no global connection; NO-GCM: the spatial extraction modules such as GCN and CLA are removed, and the model can only extract temporal features; NO-MTCN: the unified temporal features extraction module is removed; NO-STCN: the independent temporal features extraction module is removed; NO-MSTCN:  the temporal features extraction module is removed.

\begin{figure*}
	\centering
	\subfloat[RMSE and MAE errors on NYC Citi Bike]{
		\includegraphics[width=0.32\linewidth]{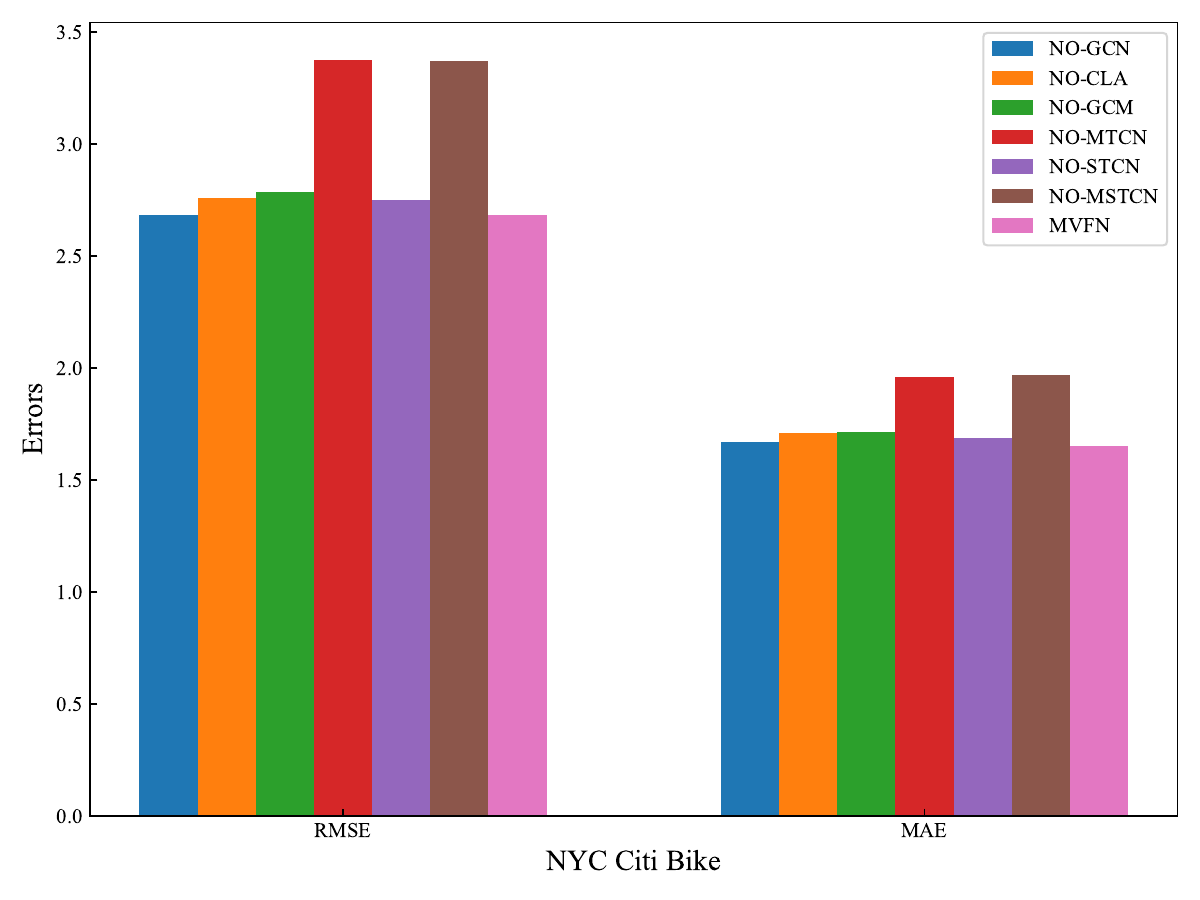}
		\label{fig: Ablation Bike}}
	\subfloat[RMSE and MAE errors on NYC Taxi]{
		\includegraphics[width=0.32\linewidth]{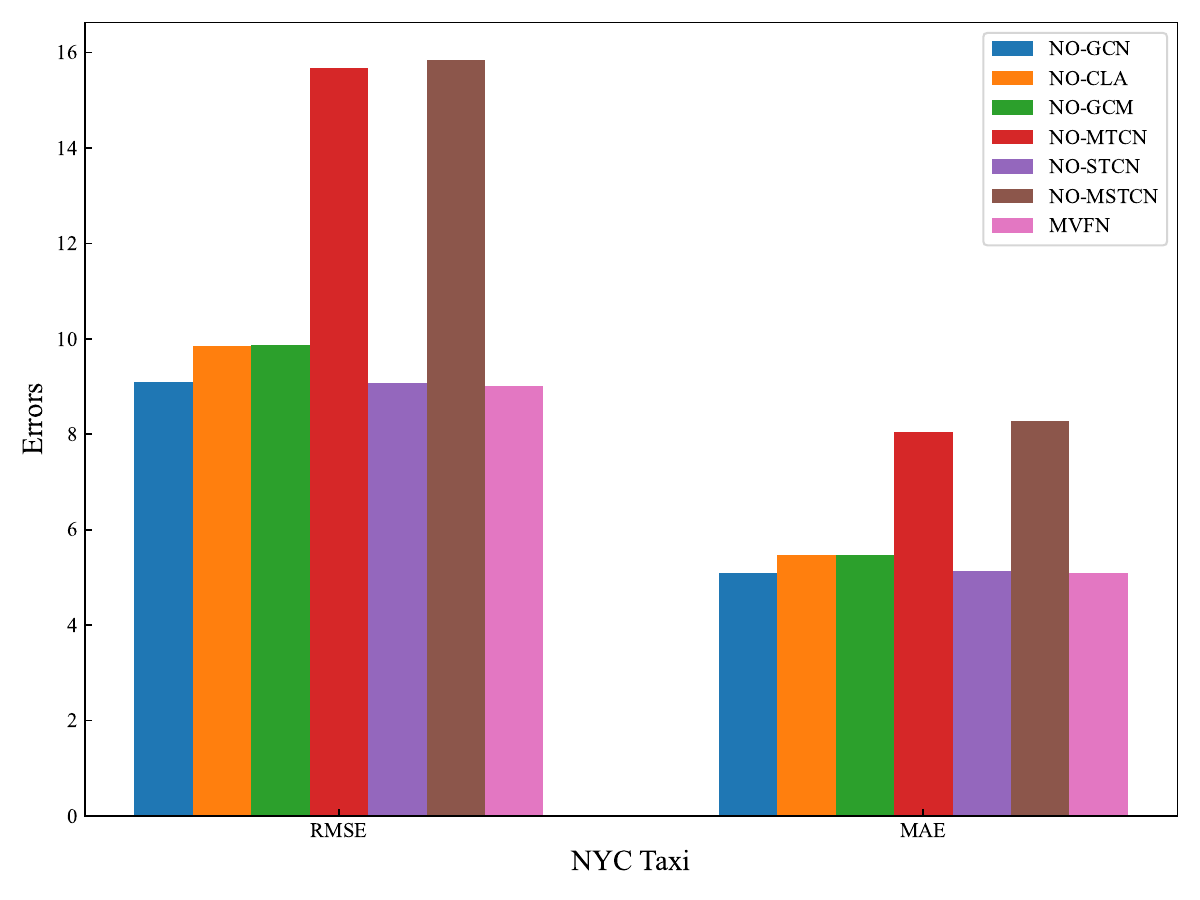}
		\label{fig: Ablation Taxi}}
	\subfloat[PCC values of different datasets]{
		\includegraphics[width=0.32\linewidth]{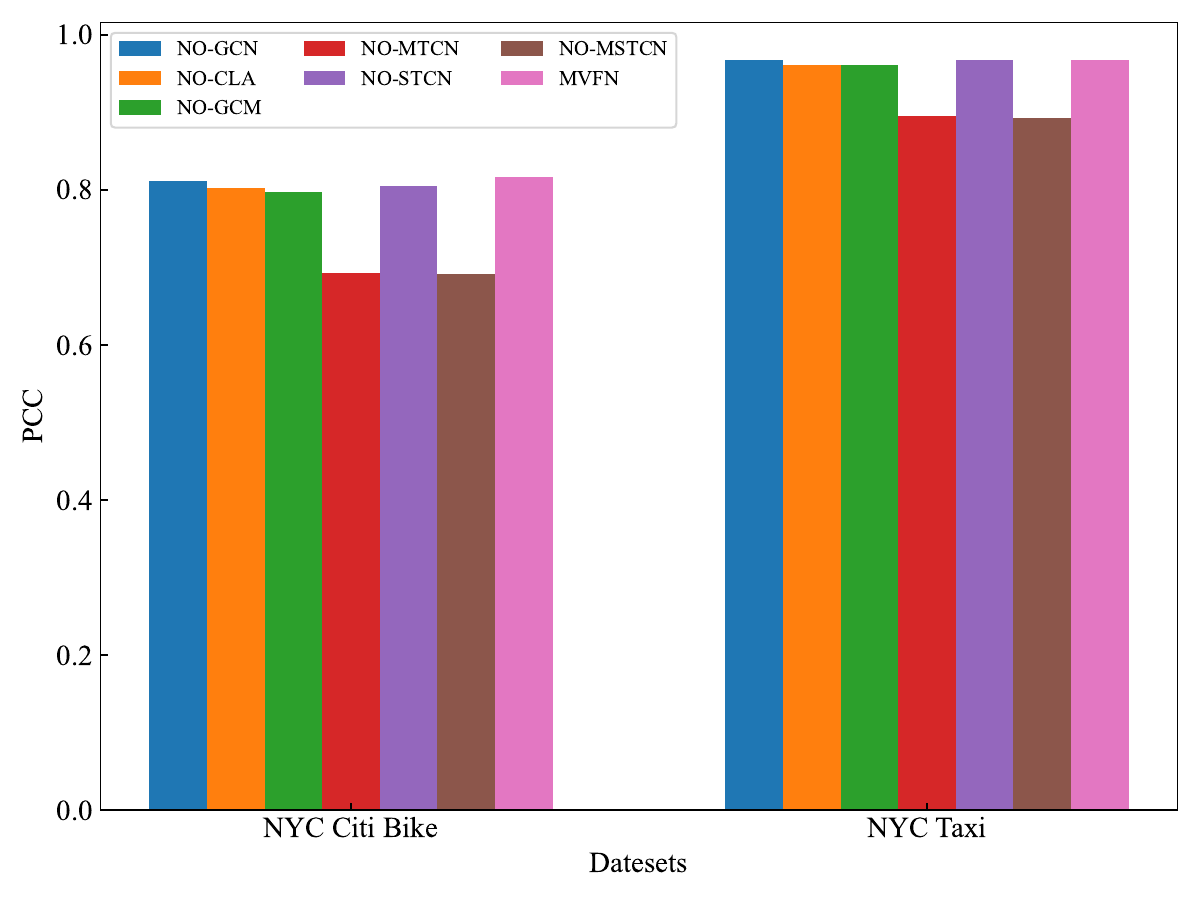}
		\label{fig: Ablation PCC}}
	
	\label{fig: Ablation}
	\caption{Ablation experiment comparison. (a): RMSE and MAE performance of each module on Bike datasets; (b): RMSE and MAE performance of each module on Taxi datasets; (c) PCC performance of each module on both datasets.}
\end{figure*}

As shown in \ref{tab:Ablation}, MVFN showed the best performance. The RMSE and MAE of the model on the NYC Citi Bike are shown in \ref{fig: Ablation Bike}, the RMSE and MAE on the NYC Taxi are shown in \ref{fig: Ablation Taxi}, and the PCC metrics of the two datasets are shown in \ref{fig: Ablation PCC}. MVFN modules of interest show similar utility on both datasets, so we only perform module effect analysis on NYC Citi Bike. The ablation experiments are analyzed with four targets, namely the spatial-temporal module, spatial module, temporal module, and the number of ST-Layers.

\begin{table}[H]
	\centering
	\renewcommand{\arraystretch}{1.2}
	\caption{Ablation experiments with different components.}
	\begin{tabular}{ccccc}
		\hline
		Datasets & Method & RMSE & MAE & PCC  \\
		\hline
		\multirow{7}{*}{NYC Citi Bike} 
		& NO-GCN     & 2.6841 & 1.6699 & 0.8115  \\
		& NO-CLA    & 2.7589 & 1.7116 & 0.8029  \\
		& NO-GCM     & 2.7867 & 1.7128 & 0.7973  \\
		& NO-MTCN    & 3.3742 & 1.9589 & 0.6923  \\
		& NO-STCN    & 2.7501 & 1.6860 & 0.8046  \\
		& NO-MSTCN   & 3.3706 & 1.9669 & 0.6919  \\
		& \textbf{MVFN }      & \textbf{2.6832} &\textbf{ 1.6504} &\textbf{ 0.8166}  \\
		\hline
		\multirow{7}{*}{NYC Taxi} 
		& No GCN     & 9.0986 & 5.0978 & 0.9671  \\
		& No CLA    & 9.8398 & 5.4660 & 0.9612  \\
		& No GCM     & 9.8701 & 5.4798 & 0.9610  \\
		& No MTCN    & 15.6682& 8.0564 & 0.8955  \\
		& No STCN    & 9.0718 & 5.1391 & 0.9671  \\
		& No MSTCN   & 15.8365& 8.2832 & 0.8921  \\
		&\textbf{ MVFN }      &\textbf{ 9.0207} & \textbf{5.0853} & \textbf{0.9674 } \\
		\hline
	\end{tabular}%
	\label{tab:Ablation}%
\end{table}%

\subsubsection{Spatial-Temporal module analysis}
\label{Spatial-Temporal module analysis}

The GCM module in MVFN is used to extract spatial features, and the MSTCN module is used to extract temporal features. Compared with MVFN, the RMSE and MAE of NO-GCM increased by 3.86\% and 3.78\%, respectively, and the PCC decreased by 2.36\%; the RMSE and MAE of NO-MSTCN increased by 25.62\% and 19.18\%, respectively, and the PCC decreased by 15.27\%. It can be seen that both the spatial and the temporal modules help to improve the accuracy of the model, and the temporal module plays a greater role.

\begin{figure*}
	\centering
	\subfloat[Error of NYC Citi Bike at different layers]{
		\includegraphics[width=0.32\linewidth]{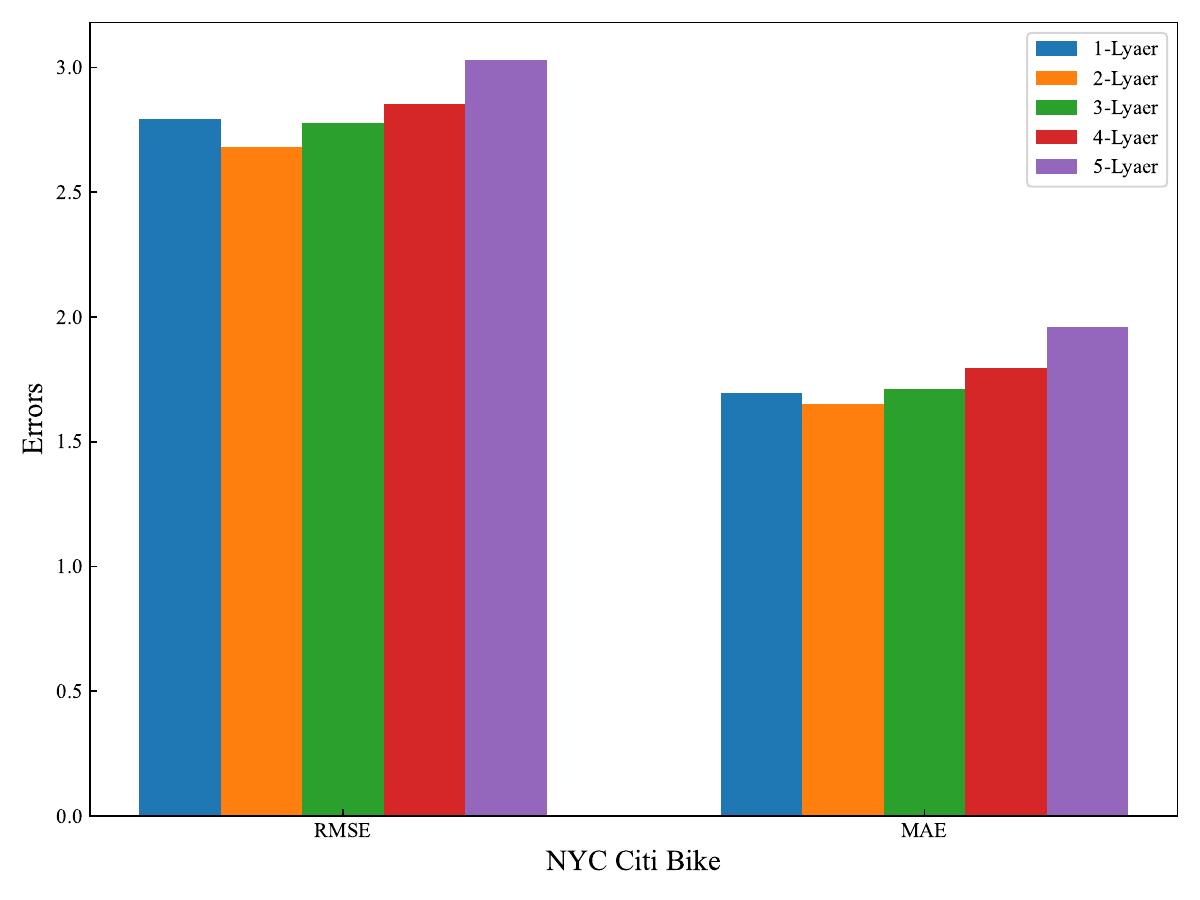}
		\label{fig: Layer Bike}}
	\subfloat[Error of NYC Taxi at different layers]{
		\includegraphics[width=0.32\linewidth]{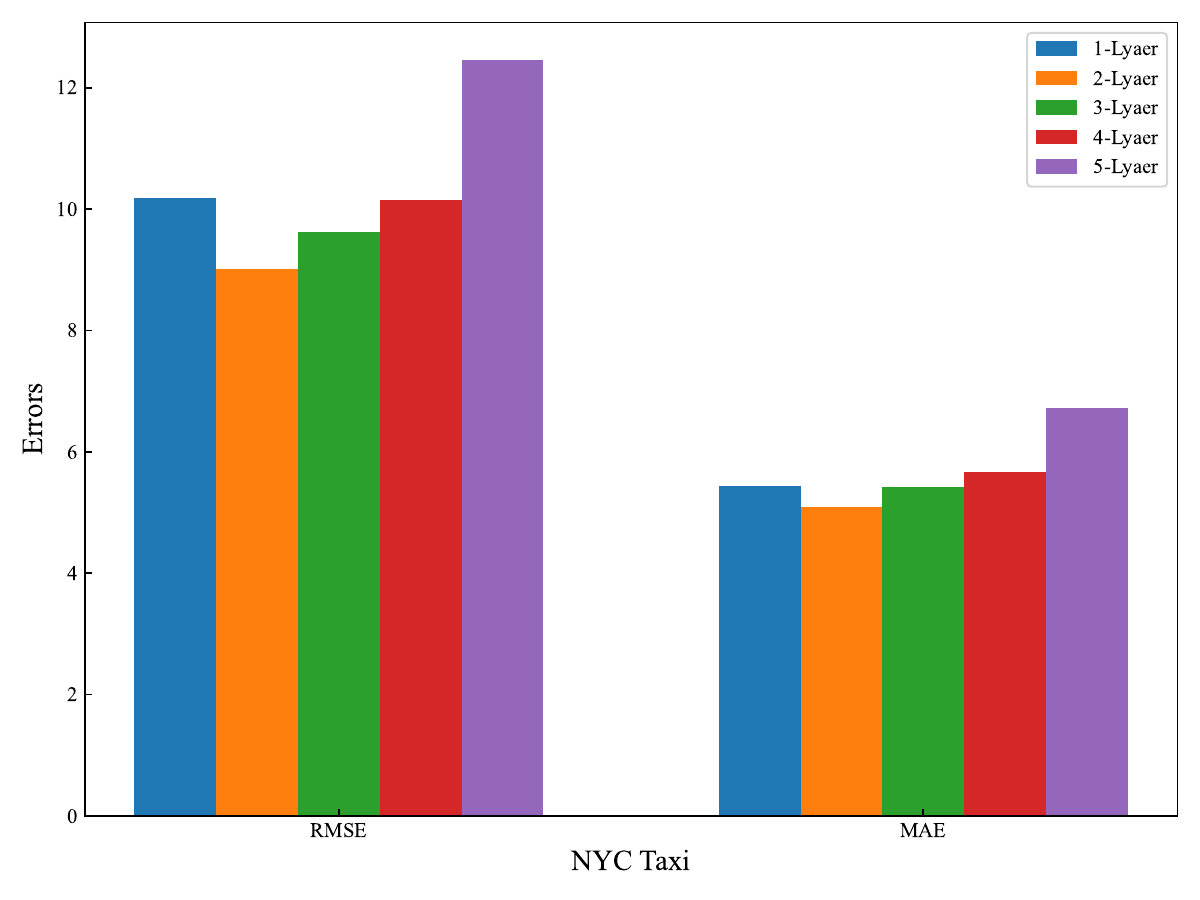}
		\label{fig: Layer Taxi}}
	\subfloat[PCC values of different layers]{
		\includegraphics[width=0.32\linewidth]{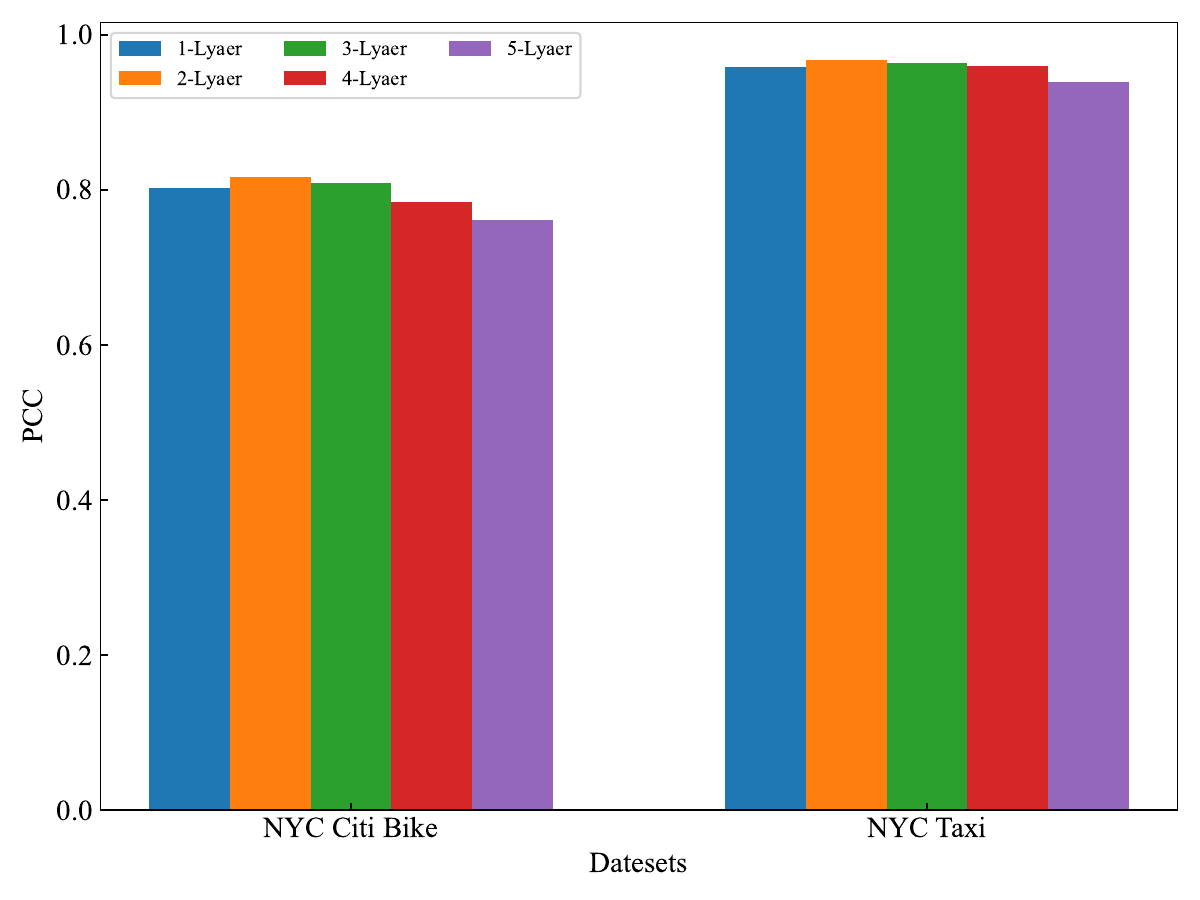}
		\label{fig: Layer PCC}}
	
	\label{fig: Layer ablation}
	\caption{Comparison of indicators for different stacking layers. (a): RMSE and MAE performance of different layers on the Bike datasets; (b): RMSE and MAE performance of different layers on the Taxi datasets; (c) Comparison of PCC metrics on different layers of the two datasets.}
\end{figure*}

\subsubsection{Spatial module analysis}
\label{Spatial module analysis}

The spatial module of MVFN is GCM, which includes two sub-modules, GCN and CLA. Compared with MVFN, the RMSE, and MAE of NO-GCN increased by 0.04\% and 1.18\%, and the PCC decreased by 0.63\%; the RMSE and MAE of NO-CLA increased by 2.82\% and 3.71\%, and the PCC decreased by 1.69\%. Both GCN and CLA contribute to the spatial features extraction of the model, and the module of CLA has a greater impact on the spatial features extraction.

\subsubsection{Temporal module analysis}
\label{Temporal module analysis}

The temporal module of MVFN is MSTCN, which includes two sub-modules, MTCN and STCN. Compared with MVFN, the RMSE, and MAE of NO-MTCN increased by 25.75\% and 18.69\%, respectively, and the PCC decreased by 15.23\%; the RMSE and MAE of NO-STCN increased by 2.5\% and 2.16\%, respectively, and the PCC decreased by 1.48\%. Both MTCN and STCN facilitate the temporal features extraction of the model, and the MTCN module has a greater impact on temporal features extraction because it can perform mutual operations on all channel of data, which plays a role in noise reduction.

\begin{table}[H]
	\centering
	\renewcommand{\arraystretch}{1.2}
	\caption{Ablation experiments with different layers.}
	\begin{tabular}{ccccc}
		\hline
		Datasets & Layer & RMSE & MAE & PCC  \\
		\hline
		\multirow{5}{*}{NYC Citi Bike} 
		& 1-Layer    & 2.7926 & 1.6966 & 0.8021  \\
		& \textbf{2-Layer}    & \textbf{2.6832} & \textbf{1.6504} & \textbf{0.8166}  \\
		& 3-Layer    & 2.7761 & 1.7132 & 0.8088  \\
		& 4-Layer    & 2.8549 & 1.7972 & 0.7848  \\
		& 5-Layer    & 3.0288 & 1.9605 & 0.7606  \\
		\hline
		\multirow{5}{*}{NYC Taxi} 
		& 1-Layer    & 10.1794 & 5.4306 & 0.9582  \\
		& \textbf{2-Layer}     & \textbf{9.0207} & \textbf{5.0853} & \textbf{0.9674}  \\
		& 3-Layer    & 9.6240  & 5.4177 & 0.9634  \\
		& 4-Layer    & 10.1560 & 5.6637 & 0.9593  \\
		& 5-Layer    & 12.4523 & 6.7171 & 0.9386  \\
		\hline
	\end{tabular}%
	\label{tab:Layer}%
\end{table}%

\subsubsection{ST-Layer number analysis}
\label{ST-Layer number analysis}

Experiments are conducted to verify the effect of the number of ST-Layers in spatial-temporal feature extraction, with variations ranging from 1 to 5 layers. The experimental results are shown in \ref{tab:Layer}. RMSE and MAE on the NYC Citi Bike are shown in \ref{fig: Layer Bike}, RMSE and MAE on the NYC Taxi are shown in \ref{fig: Layer Taxi}, and the PCC metrics of the two datasets are shown in \ref{fig: Layer PCC}. 

From the above figures and tables, it can be seen that when the number of ST-Layer is 2, the model produces the best results. one single layer cannot effectively extract spatial-temporal features, but when too many layers are stacked, it will result in smooth spatial-temporal features extraction and cannot achieve high-precision prediction.

\section{Conclusion}

In this study, a novel multi-view fusion neural network is proposed for traffic demand prediction. In the spatial dimension, fusion is implemented from local and global views, respectively. A graph convolutional network is utilized to extract spatial local features, and a cosine re-weighting linear attention mechanism is employed to extract spatial global features, which are then fused to obtain overall spatial features. In the temporal dimension, fusion is conducted from unified and independent views, respectively. A multi-channel temporal convolutional network is utilized to extract unified temporal features, while a separable temporal convolutional network is utilized to extract independent temporal features. The two networks are combined to form the multi-channel separable temporal convolutional network for the extraction of the overall temporal features. The model has been evaluated on two public datasets and achieved the promising results.

In future work, the efficacy of other low computational methods for global spatial feature extraction will be further validated, and also network structures that combine separable temporal convolutional network and recurrent neural network will be considered. 

\section*{Declaration of competing interest}
The authors declare that they have no known competing financial interests or personal relationships that could have appeared to influence the work reported in this paper.

\balance
\bibliographystyle{elsarticle-num} 				
\bibliography{cas-refs}

\end{document}